# On the Difficulty of Nearest Neighbor Search

**Junfeng He**                                                      jh2700@columbia.edu
Department of Electrical Engineering, Columbia University, New York, NY 10027, USA

**Sanjiv Kumar**                                                    sanjivk@google.com
Google Research, New York, NY 10011, USA

**Shih-Fu Chang**                                                   sfchang@ee.columbia.edu
Department of Electrical Engineering, Columbia University, New York, NY 10027, USA

## Abstract

Fast approximate nearest neighbor(NN) search in large databases is becoming popular. Several powerful learning-based formulations have been proposed recently. However, not much attention has been paid to a more fundamental question: how difficult is (approximate) nearest neighbor search in a given data set? And which data properties affect the difficulty of nearest neighbor search and how? This paper introduces the first concrete measure called Relative Contrast that can be used to evaluate the influence of several crucial data characteristics such as dimensionality, sparsity, and database size simultaneously in arbitrary normed metric spaces. Moreover, we present a theoretical analysis to prove how the difficulty measure (relative contrast) determines/affects the complexity of Local Sensitive Hashing, a popular approximate NN search method. Relative contrast also provides an explanation for a family of heuristic hashing algorithms with good practical performance based on PCA. Finally, we show that most of the previous works in measuring NN search meaningfulness/difficulty can be derived as special asymptotic cases for dense vectors of the proposed measure.

## 1. Introduction

Finding nearest neighbors is a key step in many machine learning algorithms such as spectral clustering, manifold learning and semi-supervised learning.

Rapidly increasing data in many domains such as the Web is posing new challenges on how to efficiently retrieve nearest neighbors of a query from massive databases. Fortunately, in most applications, it is sufficient to return *approximate* nearest neighbors of a query, which allows efficeint scalable search.

A large number of approximate Nearest Neighbor (NN) search techniques have been proposed in the last decade including hashing and tree-based methods, to name a few, (Datar et al., 2004; Liu et al., 2004; Weiss et al., 2008). However, the performance of all these techniques depends heavily on the data set characteristics. In fact, as a fundamental question, one would like to know how difficult is (approximate) NN search in a given data set. And more broadly, which data characteristics of the dataset affect the "difficulty" and how? The term "difficulty" here has two different but related meanings: in the context of NN search problem (independent of indexing methods), "difficulty" represents "meaningfulness", i.e., for a query, how differentiable is its NN point compared to other points? In the context of approximate NN search methods like tree or hashing based indexing methods, "difficulty" represents "complexity", i.e., what is the time and space complexity to guarantee to find the NN point (with a high probability)? These questions have not been paid much attention in the literature.

In terms of "meaningfulness" of NN search problem in a given dataset, most of the existing works have focused on the effect of one data property: dimensionality, that too in an asymptotic sense, showing that NN search will be meaningless when the number of dimensions goes to infinity (Beyer et al., 1999; Aggarwal et al., 2001; Francois et al., 2007). First, non-asymptotic analysis has not been discussed, i.e., when the number of dimensions is finite. Moreover, the effect of other crucial properties has not been stud-ied, for instance, the *sparsity* of data vectors. Since in many applications, high-dimensional vectors tend to be sparse, it is important to study the two data properties e.g., dimensionality and sparsity together, along with other factors such as database size and distance metric.

In terms of the complexity of approximate NN search methods like Locality Sensitive Hashing (LSH), some general bounds for (Gionis et al., 1999; Indyk & Motwani, 1998) have been presented. However, it has not been studied how the complexity of approximate NN search methods is affected by the difficulty of NN search problem on the dataset, and moreover, by various data properties like dimension, sparsity, etc.

The main contributions of this paper are:

1. We introduces a new concrete measure *Relative Contrast* for the meaningfulness/difficulty of nearest neighbor search problem in a given data set (independent of indexing methods). Unlike previous works that only provide asymptotic discussions for one or two data properties, we derive an explicitly computable function to estimate relative contrast in non-asymptotic case. It for the first time enables us to analyze how the difficulty of nearest neighbor search is affected by different data properties simultaneously, such as dimensionality, sparsity, database size, along with the norm of $L_p$ distance metric , for a given data set. (Sec. 2)

2. We provide a theoretical analysis on how the difficulty measure "relative contrast" determines the complexity of LSH, a popular approximate NN search method. This is the first work to relate the complexity of approximate NN search methods to the difficulty measure of a given dataset, allowing us to analyze how the complexity is affected by various data properties simultaneously. For practitioners' benefits, relative contrast also provides insights on how to choose parameters e.g., the number of hash tables of LSH, and a principled explanation of why PCA-based methods perform well in practice. (Sec. 3)

3. We reveal the relationship between relative contrast and previous studies on measuring NN search difficulty, and show that most existing works can be derived as special asymptotic cases for dense vectors of the proposed relative contrast. (Sec. 4)

## 2. Relative Contrast ($C_r$)

Suppose we are given a data set $X$ containing $n$ $d$-dim points, $X = \{x_i, i = 1, \ldots, n\}$, and a query $q$

where $x_i$, $q \in R^d$ are i.i.d samples from an unknown distribution $p(x)$. Further, let $D(\cdot, \cdot)$ be the distance function for the $d$-dimensional data. We focus on $L_p$ distances in this paper: $D(x, q) = (\sum_j |x^j - q^j|^p)^{1/p}$.

### 2.1. Definition

Suppose $D_{min}^q = \min_{i=1,\ldots n} D(x_i, q)$ is the distance to the nearest database sample[1], and $D_{mean}^q = E_x[D(x, q)]$ is the expected distance of a random database sample from the query $q$. We define the relative contrast for the data set $X$ for a query $q$ as : $C_r^q = \frac{D_{mean}^q}{D_{min}^q}$. It is a very intuitive measure of *separability* of the nearest neighbor of $q$ from the rest of the database points. Now, taking expectations with respect to queries, the relative contrast for the dataset $X$ is given as,

$$C_r = \frac{E_q[D_{mean}^q]}{E_q[D_{min}^q]} = \frac{D_{mean}}{D_{min}} \tag{1}$$

Intuitively, $C_r$ captures the notion of difficulty of NN search in $X$. Smaller the $C_r$, more difficult the search. If $C_r$ is close to 1, then on average a query $q$ will have almost the same distance to its nearest neighbor as that to a random point in $X$. This will imply that NN search in database $X$ is not very meaningful.

In the following sections, we derive relative contrast as a function of various important data characteristics.

### 2.2. Estimation

Suppose $x^j$ and $q^j$ are the $j^{th}$ dimensions of vectors $x$ and $q$. Let's define,

$$R_j = E_q[|x^j - q^j|^p], R = \sum_{j=1}^d R_j. \tag{2}$$

Both $R_j$ and $R$ are random variables (because $x^j$ is a random variable). Suppose each $R_j$ has finite mean and variance denoted as $\mu_j = E[R_j]$, $\sigma_j^2 = var[R_j]$. Then, the mean and variance of $R$ are given as,

$$\mu = \sum_{j=1}^d \mu_j, \quad \sigma^2 \leq \sum_{j=1}^d \sigma_j^2.$$

Here, if dimensions are independent then $\sigma^2 = \sum_j \sigma_j^2$. Without the loss of generality, we can scale the data such that the new mean $\mu'$ is 1. The variance of the scaled data, called normalized variance will be:

$$\sigma'^2 = \frac{\sigma^2}{\mu^2}. \tag{3}$$

---

[1]Without loss of generality, we assume that the query is distinct from the database samples, i.e., $D_{min}^q \neq 0$.

The normalized variance gives the spread of the distances from query to random points in the database with the mean distance fixed at 1. If the spread is small, it is harder to separate the nearest neighbor from the rest of the points. Next, we estimate the relative contrast for a given dataset as follows.

**Theorem 2.1** *If $\{R_j, j=1,...d\}$ are independent and satisfy Lindeberg's condition[2], the relative contrast can be approximated as,*

$$C_r = \frac{D_{mean}}{D_{min}} \approx \frac{1}{[1 + \phi^{-1}(\frac{1}{n} + \phi(\frac{-1}{\sigma'}))\sigma']^{\frac{1}{p}}} \quad (4)$$

*where $\phi$ is the c.d.f of standard Gaussian, $n$ is the number of database samples, $\sigma'$ is normalized standard deviation, and $p$ is the distance metric norm.*

**Proof:** *Since $R_j$ are independent and satisfy Lindeberg's condition, from central limit theorem, $R$ will be distributed as Gaussian for large enough $d$ with mean $\mu = \sum_j \mu_j$ and variance $\sigma^2 = \sum_j \sigma_j^2$. Normalizing the data by dividing by $\mu$, the new mean is $\mu' = 1$, and new variance is $\sigma'^2$ as defined in (3). Now, the probability that $R \leq \alpha$ for any $0 \leq \alpha \leq 1$ is given as*

$$P(R \leq \alpha) \approx \phi(\frac{\alpha - 1}{\sigma'}) - \phi(\frac{0 - 1}{\sigma'}), \quad (5)$$

*where $\phi$ is the c.d.f of standard Gaussian, and the second term in RHS is the correction factor since $R$ is always nonnegative.*

*Let's denote the number of samples for which $R \leq \alpha$ as $N(\alpha)$. Clearly, $N(\alpha)$ follows Binomial distribution with probability of success given in (5):*

$$P(N(\alpha) = k) = \binom{n}{k}(P(R \leq \alpha))^k(1 - P(R \leq \alpha))^{n-k}.$$

*Hence the expected number of database points, $\bar{N}(\alpha)$ that satisfy $R \leq \alpha$ can be computed as*

$$\bar{N}(\alpha) = E[N(\alpha)] = nP(R \leq \alpha) = n(\phi(\frac{\alpha - 1}{\sigma'}) - \phi(\frac{-1}{\sigma'})).$$

*Recall $D_{min}$ is the expected distance to the nearest neighbor and $R_{min} \approx D_{min}^p$.[3] Thus, $\bar{N}(D_{min}^p) \approx \bar{N}(R_{min}) = 1$. Hence,*

$$D_{min} \approx (\bar{N}^{-1}(1))^{\frac{1}{p}} \approx [1 + \phi^{-1}(\frac{1}{n} + \phi(\frac{-1}{\sigma'}))\sigma']^{\frac{1}{p}} \quad (6)$$

*Moreover, after normalization, $R$ follows a Gaussian distribution with mean 1. So, $R_{mean} = 1$, and $D_{mean} \approx R_{mean}^{\frac{1}{p}} = 1$. Thus, the relative contrast can be approximated as:*

$$C_r = \frac{D_{mean}}{D_{min}} \approx \frac{1}{[1 + \phi^{-1}(\frac{1}{n} + \phi(\frac{-1}{\sigma'}))\sigma']^{\frac{1}{p}}}$$

*which completes the proof.*

**Range of $C_r$:** Note that when $n$ is large enough $\phi(\frac{-1}{\sigma'}) \leq \frac{1}{n} + \phi(\frac{-1}{\sigma'}) \leq \frac{1}{2}$, so $0 \leq 1 + \phi^{-1}(\frac{1}{n} + \phi(\frac{-1}{\sigma'}))\sigma' \leq 1$ and hence $C_r$ is always $\geq 1$. And moreover, when $\sigma' \to 0$, $\phi(\frac{-1}{\sigma'}) \to 0$, and $C_r \to 1$.

**Generalization 1**: The concept of relative contrast can be extended easily to the k-nearest neighbor setting by defining $C_r^k = \frac{D_{mean}}{D_{knn}}$, where $D_{knn}$ is the expected distance to the $k^{th}$ nearest neighbor. Using $\bar{N}(D_{knn}^p) \approx \bar{N}(R_{knn}) = k$, and following similar arguments as above, one can easily show that

$$C_r^k = \frac{D_{mean}}{D_{knn}} \approx \frac{1}{[1 + \phi^{-1}(\frac{k}{n} + \phi(\frac{-1}{\sigma'}))\sigma']^{\frac{1}{p}}} \quad (7)$$

### 2.3. Effect of normalized variance $\sigma'$ on $C_r$

From (4), relative contrast is a function of database size $n$, normalized variance $\sigma'^2$, and distance metric norm $p$. Here, $\sigma'$ is a function of data characteristics such as dimensionality and sparsity. Figure 1 shows how $C_r$ changes with $\sigma'$ according to (4) when $n$ is varied from 100 to 100M, and $0 < \sigma' < 0.2$ (Note that $\sigma'$ is usually very small for high dimensional data, e.g., far smaller than 0.1). It is clear that smaller $\sigma'$ leads to smaller relative contrast, i.e., more difficult nearest neighbor search.

In the above plots, $p$ was fixed to be 1 but other values yield similar results. An interesting thing to note is that as the database size $n$ increases, relative contrast increases. In other words, nearest neighbor search is more meaningful for a larger database.[4] However, this effect is not very pronounced for smaller values of $\sigma'$.

### 2.4. Data Properties vs $\sigma'$

Since we already know the relationship between $C_r$ and $\sigma'$, by analyzing how data properties affect $\sigma'$, we will find out how data properties affect $C_r$, i.e., the difficulty of NN search. Though many data properties can be studied, in this work we focus on sparsity (a very important property in many domains involving, say, text, images and videos), together with other properties like data dimension and metric.

---

[2] Lindeberg's condition is a sufficient condition for central limit theorem to be applicable even when variables are not identically distributed. Intuitively speaking, the Lindeberg condition guarantees that no $R_j$ dominates $R$.

[3] The approximation becomes exact when metric $L_1$ is considered. For other norms (e.g., $p = 2$), bounds on $D_{min}$ can be further derived.

[4] It should not be confused with computational ease since computationally search costs more in larger databases.

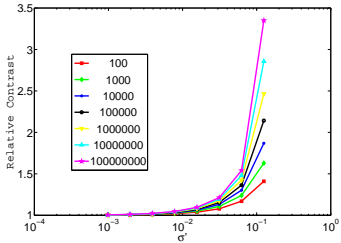

*Figure 1.* Change in relative contrast with respect to normalized data variance $\sigma'$ as in (4). The database size $n$ varies from 100 to 100M and $p = 1$. Graph is best viewed with color.

Suppose, the $j^{th}$ dimensions of vectors $x$ and $q$ are distributed the same way as a random variable $V_j$. But each dimension has only $s_j$ probability of having a non-zero value where $0 < s_j \leq 1$. Denote $m_{j,p}$ as the $p$-th moment of $|V_j|$, and $m'_{j,p}$ as the $p$-th moment of $|V_{j1} - V_{j2}|$, where $V_{j1}$ and $V_{j2}$ are independently distributed as $V_j$.

**Theorem 2.2** *If dimensions are independent,*

$\sigma'^2 = \frac{\sum_{j=1}^{d} s_j^2 m'_{j,2p} + 2(1-s_j)s_j m_{j,2p} - \mu_j^2}{(\sum_{j=1}^{d} \mu_j)^2}$

*where* $\mu_j = s_j^2 m'_{j,p} + 2(1-s_j)s_j m_{j,p}$. *Moreover, if dimensions are i.i.d.,*

$$\sigma' = \frac{1}{d^{1/2}} \sqrt{\frac{s[(m'_{2p} - 2m_{2p})s + 2m_{2p}]}{s^2[(m'_p - 2m_p)s + 2m_p]^2}} - 1. \quad (8)$$

**Proof:** *Please see the supplementary material (He, 2012).*

For some distributions, $m_p$ and $m'_p$ have a closed form representation. For example, if every dimension follows uniform distribution $U(0,1)$, then $p^{th}$ moment is quite easy in this case: $m_p = \frac{1}{(p+1)}, m'_p = \frac{2}{p+1} - \frac{2}{p+2}$. However, if $m_p$ and $m'_p$ do not have a closed form representation, one can always generate samples according to the distribution, and estimate $m_p$ and $m'_p$ empirically.

### 2.5. Data Properties vs Relative Contrast $C_r$

We now summarize how different database properties and distance metric affect relative contrast.

**Data Dimensionality** ($d$): From (8), it is easy to see that larger $d$ will lead to smaller $\sigma'$. Moreover, from (4), smaller $\sigma'$ implies smaller relative contrast $C_r$, making NN search less meaningful. This indicates the well-known phenomenon of distance concentration in high dimensional spaces. However, when dimensions are not independent, thankfully, the rate at which distances start concentrating slows down.

**Data Sparsity** ($s$): From (8), we can see that $\sigma' = \frac{1}{d^{1/2}} \sqrt{\frac{(m'_{2p} - 2m_{2p}) + \frac{2m_{2p}}{s}}{[(m'_p - 2m_p)s + 2m_p]^2}} - 1$. If $m'_p - 2m_p \geq 0$, when $s$ becomes smaller (i.e., data vectors have fewer non-zero elements), $\sigma'$ gets larger, and so does the relative contrast. Another interesting case is when $p \to 0_+$, i.e., $L_0$ or zero-one distance. In this case, $m_p = m'_p = 1$, and from (8) $\sigma' = \frac{1}{d^{1/2}} \sqrt{\frac{(1-s)}{1-(1-s)^2}}$, which increases monotonically as $s$ decreases. However, for general cases, it is not easy to theoretically prove how $\sigma'$ will change when $s$ gets smaller. But in experiments, we have always found that smaller $s$ will lead to larger $\sigma'$. In other words, when data vectors become more sparse, NN search becomes easier. That raises another interesting question: What is the effective dimensionality of sparse vectors? One may be tempted to use $d \cdot s$ as the intrinsic dimensionality. But as we will show in the experimental section, this is generally not the case and relative contrast provides an empirical approach to finding intrinsic dimensionality of high-dimensional sparse vectors.

**Database Size** ($n$): From (4), keeping $\sigma'$ fixed, $C_r$ increases monotonically with $n$. Hence, NN search is more meaningful in larger databases. Actually, when $n \to \infty$, irrespective of $\sigma'$, $1 + \phi^{-1}(\frac{1}{n} + \phi(\frac{-1}{\sigma'}))\sigma' \to 0$, and $C_r \to \infty$. Thus, when the database size is large enough, one doesn't need to worry about the meaningfulness of NN search irrespective of the dimensionality. However, unfortunately when dimensionality is high, $C_r$ increases very slowly with $n$, making the gains not very pronounced in practice. This is the same phenomenon noticed in Fig. 1 for small values of $\sigma'$.

**Distance Metric Norm** ($p$): Since $p$ appears in both (4) and (8), it makes analysis of relative contrast with respect to $p$ not as straightforward. In the special case when data vectors are dense (i.e., $s = 1$), and each dimension is i.i.d with uniform distribution, one can show that smaller $p$ leads to bigger contrast.

### 2.6. Validation of Relative Contrast

To verify the form of relative contrast derived in Sec. 2, we conducted experiments with both synthetic and real-world datasets, which are summarized below.

#### 2.6.1. SYNTHETIC DATA

We generated synthetic data by assuming each dimension to be i.i.d from uniform distribution $U[0,1]$. Fig.

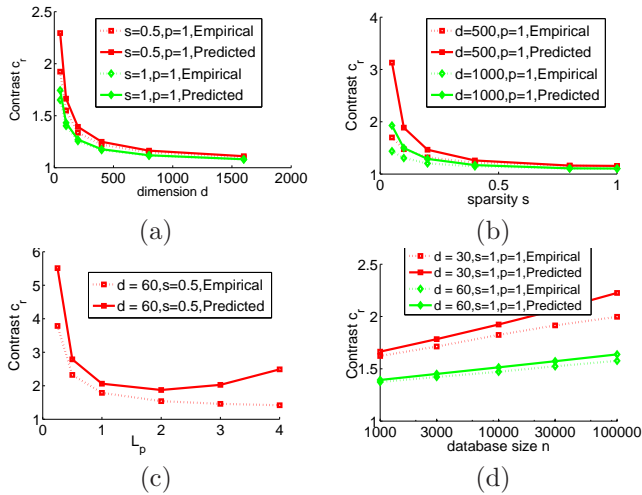

*Figure 2.* Experiments with synthetic data on how relative contrast changes with different database characteristics. Graphs are best viewed with color.

2 compares the predicted (theoretical) relative contrast with the empirical one. The solid curves show the predicted contrast computed using (4), where the normalized variance $\sigma'$ is estimated using (8). The dotted curves show the empirical contrast, directly computed according to the definition in (1) from the data by averaging the results over one hundred queries. For most of the cases, the predicted and empirical contrasts have similar values.

Fig. 2 (a) confirms that as dimensionality increases, relative contrast decreases, thus making the nearest neighbor search harder. Moreover, except for very small $d$, the prediction is close to the empirical contrast verifying the theory. It is not surprising that predictions are not very accurate for small $d$ since the central limit theorem(CLT) is not applicable in that case. It is interesting to note that (4) also predicts the rate at which contrast changes with $d$, unlike the previous work (Beyer et al., 1999; Aggarwal et al., 2001) which only show that NN search becomes impossible when dimensionality goes to infinity.

Fig. 2 (b) shows how data sparsity affects the contrast for two different choices of $d$. The main observation is that as $s$ increases (denser vectors), contrast decreases, making nearest neighbor search harder. In other words, lesser the number of non-zero dimensions for a fixed $d$, easier the search. In fact, the search remains well-behaved even in high-dimensional datasets if data is sparse. The prediction is quite accurate in comparison to the empirical one except when $s.d$ is small and hence CLT does not apply any more. As a note of caution, one should not regard $s.d$ as the intrinsic dimensionality of the data, since a dataset with dense vectors of dimension $s.d$ usually has different

*Table 1.* Description of the real-world datasets. $n$ - database size, $d$ - dimensionality, $s$ - sparsity (fraction of nonzero dimensions), $d_e$ - effective dimensionality containing 85% of data variance.

|  | n | d | s | $d_e$ |
|---|---|---|---|---|
| gist | 95000 | 384 | 1 | 71 |
| sift | 95000 | 128 | 0.89 | 40 |
| color (histograms) | 95000 | 1382 | 0.027 | 22 |
| image (bag-of-words) | 95000 | 10000 | 0.024 | 71 |

contrast than the $d$-dim $s$-sparse data set.

The effects of two other characteristics i.e., $L_p$ distance metric for different $p$ and database size $n$ are shown in Figs. 2 (c) and (d), respectively. The effect of these parameters on relative contrast is milder than that of $d$ and $s$. For large $d$, the contrast drops quickly and it becomes hard to visualize the effects of $p$ and $n$. So, here we show these plots for smaller values of $d$. From Fig. 2 (c) it is clear that for norms less than 1, contrast is the highest (Note that we have an approximation for $p > 1$ in Theorem 2.1, which causes the bias of predicted $C_r$ for $p = 3, 4$). This observation matches the conclusion from (Aggarwal et al., 2001) for dense vectors. Fig. 2 (d) shows that as the database size increases, it becomes more meaningful to do nearest neighbor search. But as the dimensionality is increased (from 30 to 60 in the plot), the rate of increase of contrast with $n$ decreases. For very high dimensional data, the effect of $n$ is very small.

### 2.6.2. REAL-WORLD DATA

Next, we conducted experiments with four real-world datasets commonly used in computer vision applications: *sift, gist, color* and *image*. The details of these sets are given in Table 1. The sift and gist sets contain 128-dim and 384-dim vectors, which are mostly dense. On the other hand, both color and image datasets are very high dimensional as well as sparse. Color data set contains color histogram of images while the image data set contains bag-of-words representation of local features in images.

While deriving the form of relative contrast in Sec. 2, we assumed that dimensions were independent. However, this assumption may not be true for real-world data. One way to adress this problem would be to assume that the dimensions become independent after embedding the data in an appropriate low-dimensional space. In these experiments, we define effective dimensionality $d_e$ as the number of dimensions necessary to preserve 85% variance of the data[5]. The effective dimensionality for different datasets is shown in Table

---

[5] For large databases, one can use a small subset to estimate the covariance matrix.

*Table 2.* Experiments with four real-world datasets. Here, predicted contrast is computed using the effective dimensionality containing 85% of data variance.

| | p=1 | p=2 |
|---|---|---|
| gist empirical contrast | 1.83 | 1.78 |
| gist predicted contrast | 1.62 | 1.87 |
| sift empirical contrast | 4.78 | 4.23 |
| sift predicted contrast | 2.03 | 3.94 |
| color empirical contrast | 3.19 | 4.81 |
| color predicted contrast | 2.78 | 8.10 |
| image empirical contrast | 1.90 | 1.66 |
| image predicted contrast | 1.62 | 1.87 |

1. Table 2 compares the empirical and predicted relative contrasts for different datasets. Since our theory is based on the law of large numbers, the prediction is more accurate on image and gist data sets as their effective dimensions are large enough. For the color data, $d_e$ is too small (just 22) and hence the prediction of relative contrast shows more bias for this set.

One interesting outcome of these experiments is that our analysis provides an alternative way of finding intrinsic dimensionality of the data which can be further used by various nearest neighbor search methods. The traditional method of finding intrinsic dimensionality using data variance suffers from the assumption of linearity of the low-dimensional space and the arbitrary choice of threshold on variance. On the other hand, nonlinear methods are computationally prohibitive for large datasets. In the relative contrast based method, for a given dataset, one can sweep over different values of $d'$ where $0 < d' < d$, and find the one which gives the least discrepancy between the predicted and empirical contrasts averaged over different p. For large datasets, one can use a smaller sample and a few queries to estimate the empirical contrast. Using this procedure, the intrinsic dimensionality for the four datasets turns out to be: sift - 41, gist - 75, color - 41, image - 70. For the two sparse datasets (color and image), it indicates the dimensionality of equivalent low-dimensional *dense* vector space. It is interesting to note that intrinsic dimensionality is not equal to $d \cdot s$ for the two sparse datasets as discussed before. For image dataset, it is much smaller than $d \cdot s$ indicating high correlations in non-zero entries of the data vectors.

## 3. Relative Contrast and Hashing

### 3.1. Relative Contrast and LSH

LSH are commonly used in many practical large-scale search systems due to their efficiency and ability to deal with high-dimensional data. In each hash table, every data point $x$ is converted into codes by using a series of $k$ hash functions $h_j(x), j = 1, \cdots, k$. Each hash function is designed to satisfy the locality condition i.e., neighboring points have the same hashed value with high probability and vice-a-versa. A commonly used hash function in LSH is $h(x) = \lfloor \frac{w^T x + b}{t} \rfloor$, where $w$ is a vector with entries sampled from a $p$-stable distribution, and $b$ is uniformly distributed as $U[0, t]$ (Datar et al., 2004). We now provide the following theorems to show how relative contrast ($C_r$) affects the complexity of LSH.

**Theorem 3.1** *LSH can find the exact nearest neighbor with probability $1 - \delta$ by returning $O(\log \frac{1}{\delta} n^{g(C_r)})$ candidate points, where $g(C_r)$ is a function monotonically decreasing with $C_r$.*

**Proof:** Please see the supplementary material.

**Corollary 3.2** *LSH can find the exact nearest neighbor with a probability at least $1 - \delta$ with a time complexity $O(d \log \frac{1}{\delta} n^{g(C_r)} \log n)$ and space complexity $O(\log \frac{1}{\delta} n^{(1+g(C_r))} + nd)$. $l$, the number of hash tables needed, is $l = O(\log \frac{1}{\delta} n^{g(C_r)})$.*

**Proof:** Please see the supplementary material.

The above theorems imply that when $C_r$ is larger, $g(C_r)$ will be smaller, thus, among the datasets of same size, to get the same recall of the true nearest neighbor, the dataset with higher relative contrast $C_r$ will have better time and space complexity, return less number of candidates for reranking, and need fewer number of hash tables, or in one word, be easier for approximate NN search with LSH.

Note that our theory shares some similarity to the results in (Gionis et al., 1999) about the complexity of LSH, however, it has several unique properties. First, our theory is about finding exact NN (with a probability guarantee), not finding approximate NN (with a probability guarantee) like in previous works. Moreover, we have related the complexity of LSH to relative contrast $C_r$, enabling us to analyze how the complexity of LSH is affected by various data properties of the dataset simultaneously. To the best of our knowledge, our work is the first one on this important topic.

To verify the effect of relative contrast on LSH, we conducted experiments on three real-world datasets.

In Fig. 3, performance of LSH for $L_1$ distance (i.e., $p = 1$) is given on three datasets: sift, gist and color. From Table 2, for $p = 1$, $C_r$ for the three datasets is in this order: sift(4.78) > color(3.19) > gist (1.83). From Fig. 3 (a), we can see that for several settings of number of bits and number of tables, the number of returned points needed to get the same nearest neighbor recall for the three sets follows sift < color < gist, as predicted by Theorem 3.1. Moreover, from Fig. 3 (b),

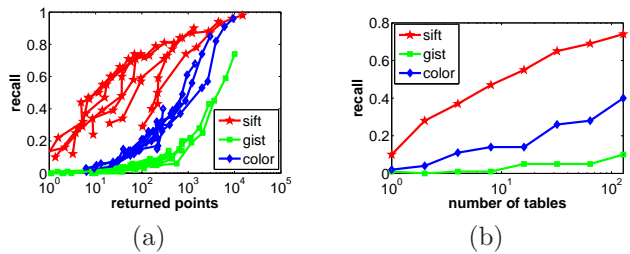

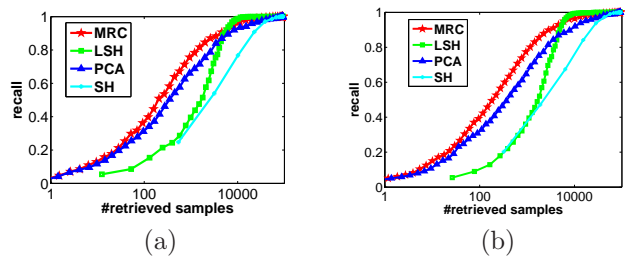

*Figure 3.* Performance of LSH on three datasets: sift, gist, and color. (a) Recall of the nearest neighbor. Each curve represents different number of bits, e.g., $k = 12, 16, ...40$. Each marker on the curve represents different number of hash tables $l$, e.g., $l = 1, 2, ...128$. (b) Recall of the nearest neighbor for different number of hash tables for $k = 32$. Graphs are best viewed with color.

*Figure 5.* Recall of 1-NN for hamming reranking with different hashing methods on color data using (a) 80 bits, (b)100 bits. Relative contrast based method (MRC) can improve upon PCA-based hashing. Graphs are best viewed with color.

commonly used hash function in PCA-based hashing methods is

$$h(x) = sgn(w^T x + b) \qquad (9)$$

where $w$ is heuristically picked as a PCA direction, and $b$ is a threshold which is usually chosen as $E[w^T x]$. Assuming the data to be zero-centered, i.e., $E[x] = 0$, leads to $b = 0$. Since $q$ and $x$ are assumed to be i.i.d samples from some unknown $p(x)$, $E[q] = 0$ as well.

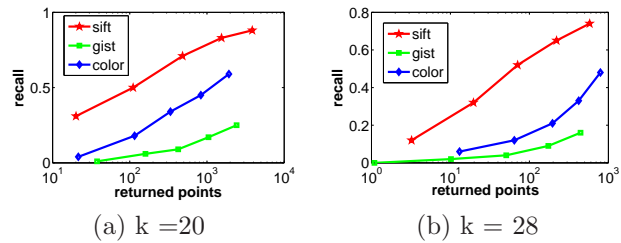

*Figure 4.* Recall vs the number of returned points when using hamming ranking. Number of bits $k = 20$ for (a) and $k = 28$ for (b). Graphs are best viewed with color.

the number of hash tables needed to get the same recall follows sift < color < gist, as predicted by Corollary 3.2. We have tried experiments with $k = 12, 16..., 40$ and observe the same trend, but only show results for $k = 32$ due to space limit.

The above experiments used the typical framework of hash table lookup. Another popular way to retrieve neighbors in code space is via hamming ranking. When using a $k$-bit code, points that are within hamming distance $r$ to the query are returned as candidates. In Figure 4, we show the recall of nearest neighbor for two different values of $k$. Similar to the case of hash table lookup experiments, the number of returned points needed to get the same recall follows sift < color < gist. This follows the same order as suggested by relative contrast. The interesting thing is that color has much higher dimensionality than gist, but its sparsity helps in achieving better relative contrast and hence better search performance.

### 3.2. Relative Contrast and PCA hashing

Hashing methods that use PCA as a heuristics often achieve quite good performance in practice (Weiss et al., 2008; Gong & Lazebnik, 2011). In this section, we show PCA hashing is actually seeking projections that maximize relative contrast in each projection with $L_2$ distance under some assumptions. A

For a query $q$, denote $x_{q,NN}$ as q's NN in the database. Denote $S_{NN} = E_q[(q - x_{q,NN})(q - x_{q,NN})^T]$, and $\Sigma_X = (1/n) \sum_i x_i x_i^T$. The following theorem shows that maximizing relative contrast will lead us to PCA hashing under some assumptions.

**Theorem 3.3** *For linear hashing as (9), to find projection vector $w$ to maximize relative contrast, we should find $\hat{w} = \arg\max_w \frac{w^T \Sigma_X w}{w^T S_{NN} w}$. If we further assume that the nearest neighbors are isotropic, i.e., $S_{NN} = \alpha I$, we will get $\hat{w} = \arg\max_w w^T \Sigma_X w$, i.e., PCA hashing.*

**Proof:** Please see the supplementary material.

If we do not assume nearest neighbors to be isotropic, we can empirically compute $S_{NN}$ from a few samples. And then we can find projection vectors $w$ in (9) as $\hat{w} = \arg\max_w \frac{w^T \Sigma_X w}{w^T S_{NN} w}$, which are the generalized eigenvectors of $\Sigma_X$ and $S_{NN}$. This will often obtain better results than PCA hashing. We provide one example in Figure 5, in which "MRC" represents the method we described as above, and "PCA", "LSH", "SH" are PCA hashing, Locality Sensitive Hashing, and Spectral Hashing (Weiss et al., 2008) respectively.

## 4. Related Works

### 4.1. Previous Works

Some of the influential works on analyzing NN search difficulty are (Beyer et al., 1999) and (Francois et al.,

2007), whose main results are shown in Theorem 4.1 and 4.2.

**Theorem 4.1** *(Beyer et al., 1999)* Denote $D_{max}^q = \max_{i=1,...n} D(x_i, q)$ and $D_{min}^q = \min_{i=1,...n} D(x_i, q)$. If $\lim_{d \to \infty} var(\frac{D(x_i,q)^p}{E[D(x_i,q)^p]}) \to 0$, then for every $\epsilon \geq 0$, $\lim_{d \to \infty} P[D_{max}^q \leq (1+\epsilon)D_{min}^q] = 1$.

**Theorem 4.2** *(Francois et al., 2007)* If every dimension of the data is i.i.d., when $d \to \infty$, $\frac{\sqrt{Var(||x_i-q||_p)}}{E(||x_i-q||_p)} \approx \frac{1}{\sqrt{d}} \frac{1}{p} \frac{\sigma_j}{\mu_j}$, where $\sigma_j = Var(||x_i^j - q^j||_p^p)$ and $\mu_j = E(||x_i^j - q^j||_p^p)$ are the variance and mean on each dimension.

### 4.2. Relations Between Our Analysis and Previous Works

**Relation to Beyer's Work**

Note that if the distance function $D(x_i, q)$ in Beyer's work is $L_p$ distance, then $var(\frac{D(x_i,q)^p}{E[D(x_i,q)^p]}) = \frac{\sigma^2}{\mu^2} = (\sigma')^2$. When $\sigma' \to 0 (d \to \infty)$, Beyer's work shows that $D_{max}^q \approx D_{min}^q$, and our theory shows $C_r \to 1$, or equivalently $D_{mean} \to D_{min}$. So we will get the same conclusion: when $d \to \infty$, NN search is not very "meaningful", because we can not differentiate the nearest neighbor from other points. However, Beyer's theory works for the worst case (i.e., compare NN point to the worst point with maximum distance), while ours works for the average case.

**Relation to Francois's Work**

In Theorem 4.2, a measurement called "relative variance", defined as $\frac{\sqrt{Var(||x_i-q||_p)}}{E(||x_i-q||_p)}$, is discussed, which is a modification of the condition $var(\frac{D(x_i,q)^p}{E[D(x_i,q)^p]})$ in Beyer's work. If $\frac{\sqrt{Var(||x_i-q||_p)}}{E(||x_i-q||_p)} \to 0$, NN search will become meaningless. The following theory reveals the relationship between relative variance and relative contrast.

**Theorem 4.3** *In (4), if $\sigma' \to 0$ (e.g., $d \to \infty$), $C_r \approx \frac{1}{1 + \phi^{-1}(\frac{1}{n})\frac{1}{d^{1/2}}\frac{\sigma_j}{\mu_j}}$.*

**Proof:** Please see the supplementary material.

From Theorem 4.3, we see when $\sigma' \to 0$ (e.g., $d \to \infty$), the relative contrast monotonically depends on $\frac{1}{p}\frac{1}{d^{1/2}}\frac{\sigma_j}{\mu_j}$, which equals to "relative variance" as in Theorem 4.2.

To summarize, most of the known analysis can be derived as special asymptotic cases (when $\sigma' \to 0$, e.g., $d \to \infty$) of the proposed measure with the focus on only one or two data properties.

## 5. Conclusion and Future Work

In this work, we introduced a new measure called relative contrast to describe the difficulty of nearest neighbor search in a data set. The proposed measure can be used to evaluate the influence of several crucial data characteristics such as dimensionality, sparsity, and database size simultaneously in arbitrary normed metric spaces. Furthermore, we show how relative contrast determines the difficulty of ANN search with LSH and provides guidance for better parameter settings. In the future, we would like to relax the independence assumption used in the theory of relative contrast, and also study how relative contrast affects the complexity of other approximate NN search methods besides LSH. Moreover, we will explore a better but harder definition of $C_r = E_q[\frac{D_{mean}^q}{D_{min}^q}]$.